%% file: strawberry_paper.tex
\documentclass{svproc}
\usepackage{url}
\usepackage{orcidlink}

\usepackage[acronym,toc]{glossaries}
\usepackage{colortbl,hhline}

\loadglsentries{glossary}
\makeglossaries

\begin{document}
\mainmatter              
\title{Deep Learning approach for Classifying Trusses and Runners of Strawberries}
\titlerunning{Deep Learning approach for Classifying ...}  
%
\author{Jakub Pomykala, Francisco de Lemos\orcidlink{0000-0003-1751-764X}, Isibor Kennedy Ihianle\orcidlink{0000-0001-7445-8573}, David Ada Adama\orcidlink{0000-0002-2650-857X}, Pedro Machado\orcidlink{0000-0003-1760-3871}}
\authorrunning{Pomykala et al.} 
%
\tocauthor{Jakub Pomykala, Francisco Lemos, Isibor Kennedy Ihianle, David Adama, Pedro Machado}
\institute{Computational Intelligence and Applications Research Group \\ Department of Computer Science\\ School of Science and Technology\\
Nottingham Trent University\\ Clifton Campus NG11 8NS Nottingham\\
\email{jakub.pomykala2017@my.ntu.ac.uk,\\ \{francisco.lemos, isibor.ihianle, david.adama,pedro.machado\}@ntu.ac.uk}}

\maketitle              

\begin{abstract}
The use of artificial intelligence in the agricultural sector has been growing at a rapid rate to automate farming activities. Emergent farming technologies focus on mapping and classification of plants, fruits, diseases, and soil types. Although, assisted harvesting and pruning applications using deep learning algorithms are in the early development stages, there is a demand for solutions to automate such processes. This paper proposes the use of Deep Learning for the classification of trusses and runners of strawberry plants using semantic segmentation and dataset augmentation. The proposed approach is based on the use of noises (i.e. Gaussian, Speckle, Poisson and Salt-and-Pepper) to artificially augment the dataset and compensate the low number of data samples and increase the overall classification performance. The results are evaluated using mean average of precision, recall and F1 score. The proposed approach achieved 91\%, 95\% and 92\% on precision, recall and F1 score, respectively, for truss detection using the ResNet101 with dataset augmentation utilising Salt-and-Pepper noise; and 83\%, 53\% and 65\% on precision, recall and F1 score, respectively, for truss detection using the ResNet50 with dataset augmentation utilising Poisson noise.

\keywords{Artificial Intelligence, Image Augmentation, Strawberry Trusses, Strawberry Runners}
\end{abstract}
\pagenumbering{gobble} 

\section{Introduction}

\gls*{ai} is commonly defined as the system’s ability to interpret external data correctly, to learn from such data, and to use the knowledge obtained to achieve specific goals and execute tasks through flexible adaptation, \cite{haenlein}. \gls*{ai} is used to perform complex tasks that would otherwise require humans to complete, leading to increased automation and precision when these tasks are completed. \gls*{dl} methods are widely used for pattern recognition by simulating the brain plasticity. \glspl*{cnn} are a type of \gls*{dl} algorithms for the classification of visual data.

Due to the fact that businesses are discovering that tasks can be automated using \gls*{ai}, the latter is becoming increasingly used to replace manual work, therefore lowering costs and minimising human error. Furthermore, with the current scarcity of seasonal employees in the United Kingdom, farmers are giving out fruits and vegetables to minimise wastes \cite{bbc2021}, 
hence the need to automate agricultural processes using \gls*{ai}. 
Strawberry cultivation is already being automated, and \gls*{ai} is being used to further automate the process. With this, it is possible to segment the images of strawberries by understanding their structure. In \cite{poling2012strawberry} the structure of a strawberry plant is detailed, as well as what each part of the structure does. Flowers and, eventually, fruits sprout from the trusses. Strawberry runners are horizontal stems that run above ground and create new clone plants at the end of the long stem. By taking the runner from the plant, farmers are able to preserve the strawberry varieties which are yielding better fruit. It is possible to detect which plants are more suited for farming using \gls*{dl} algorithms, and so boost strawberry harvest productivity.

In this paper, Semantic Segmentation is proposed to classify runners and trusses of strawberries from images and/or videos. The identification of trusses and runners is required to automate the process of auto pruning (old trusses and runners removal) and autonomous strawberry harvesting (ripping strawberries from their trusses). It is intended that the proposed methodology will help to improve farming processes and can be easily adopted to other crops. The remainder of this paper is composed of, Section \ref{sec:lr} where we perform an extensive literature review of \gls*{dl} being used in farming. Section \ref{sec:imgsegmodel} explains the \gls*{dl} model to classify strawberry trusses and runners. Section \ref{sec:resultsevaluations} presents the results obtained by the proposed model and evaluates them, and Section \ref{sec:conlusionsfw} draws conclusions and future work.

\section{Literature Review}\label{sec:lr}
\gls*{ai} is being used to increase the productivity and reduce wastes by enabling farmers to make informed decisions on farming practises \cite{zheng2021remote}. Substantial studies have been conducted using different \gls*{dl} methods and computational models for the classification of fruits and plants \cite{behera2020,naik2017,bhargava2021}.

Bargoti et al. \cite{Bargoti2017} uses the Faster \gls*{rcnn} for the classification of orchards, such as mangoes, almonds, and apples. The Faster \gls*{rcnn} framework showed an outstanding performance on multiple orchard image data. 
The use of data augmentation techniques such as flip and scale augmentations were useful to improve performance with varying number of training images, resulting in equivalent performance with less than half the number of training images.The paper presented an F1 score of 90\% for mangoes and apples classification.

Lamb et al. \cite{Lamb2018} proposed using a system based on \glspl*{cnn} for strawberry detection. Additionally, ablation studies are presented to validate the choice of the hyperparameters, framework, and network structure. The work also introduces input compression, image tiling, colour masking, and network compression to both the training data and network structure to improve precision and computational time. The implementation of the work was done using a Raspberry Pi 3B\footnote{Available online, \protect\url{https://www.raspberrypi.com/products/raspberry-pi-3-model-b/}, last accessed 17/06/2022} with a detection speed of 1.63 \gls*{fps} and an average precision of 84.2\%. Due to the system's effectiveness, unmanned robots could be used to replace humans as harvesters, potentially increasing agricultural efficiency and lowering crop damage.

A high segmentation accuracy and effective extraction of greenhouse fruit parts can be achieved using \gls*{lda}. Wang and Lihong \cite{wang2018unsupervised} performed an unsupervised segmentation of greenhouse plants based on \gls*{msbslda}. The comparison experiments show that the proposed \gls*{msbslda} algorithm achieves good expression of image detail and global structure information at the same time, allowing it to accurately include the lesion part of the plant and distinguish it from mosses of similar colour by adjusting the topic number.

Kestur et al. \cite{Kestur2019} proposed the MangoNet architecture for mango classification using semantic segmentation. The MangoNet was trained using 11096 image patches of size 200$\times$200 obtained from 40 images. Testing is carried out using 1500 image patches created from 4 test images. The results are analysed for performance of segmentation and detection of mangoes using the precision, recall and F1 score based on the contingency matrix. The authors claim that the model is robust for the mango detection, irrespective of scale, occlusion, distance and illumination conditions, characteristic to open field conditions. Whilst comparing the performance of the MangoNet with \gls*{fcn} architectures trained on the same data, MangoNet outperforms its variant architectures.

Titan et al. \cite{tian2019} proposed the detection of apples in different growth stages using YOLOv3 model. This work considers fluctuating illumination, complex backgrounds, overlapping apples and branches and leaves. The images used were augmented using rotation transformation, colour balance transformation, brightness transformation and blur processing. The authors then used a DenseNet architecture to process feature layers with low resolution in the YOLOv3 network. This effectively enhances feature propagation, improves feature reuse and network performance. The YOLOv3 is an effective \gls{cnn} architecture for the classification of apples under challenging classification scenarios (such as partial occlusion, low visibility, and severe weather conditions) and can be used in real-time applications \cite{tian2019}. 

Kang et al.\cite{Kang2020} suggested the DaSNet-v2 for instance segmentation on fruits, and semantic segmentation on branches. DaSNet-v2 uses ResNet101 achieves 86,8\%, 88\% and 87,3\% on recall, precision of detection, and accuracy of instance segmentation on fruits. The accuracy for branch segmentation is 79,4\%. DaSNet-v2 with light-weight backbone Resnet-18 achieves 85\%, 87\% and 86,6\% on recall, precision of detection, and accuracy of instance segmentation on fruits. As for accuracy of branch segmentation, the result obtained is 77,5\%. The authors claim that DaSNet-v2 can robustly and efficiently perform the vision sensing for robotic harvesting in apple orchards.

Although the focus of this paper is on strawberry farming, the proposed methodology can be applied to other areas of farming (from production to commercialisation). Identifying strawberry diseases from its leaf can prevent damage to the fruit. Some diseases can be detected from the leaf, examples of which includes spot leaf, blight leaf and scorch leaf. Ramadani et al. \cite{ramdani2021strawberry} used ResNet50 architecture with 3600 images to create a model to identify strawberry diseases from its leaf. When a strawberry runner is replanted it will grow a clone of the original strawberry plant, hence being able to detect the leaf disease allows farmers to know in advance which plants are healthy. The model achieved 99.9\% accuracy for the spot leaf, 99\% for scorch leaf and 99.9\% for a healthy leaf, therefore, showing that it is possible to identify diseases in strawberry leafs.

Emerging technology in this area supports farm tasks like yield mapping or robotic harvesting \cite{boursianis2022internet}. The latter can reduce the costs of labour and increase the overall quality of fruit produced. The research achieved higher accuracy of detection and quicker processing times in comparison to the traditional detectors. The authors in \cite{boursianis2022internet} claim that \gls*{ai} has great potential in the development of an autonomous and real time harvesting or yield mapping/estimation system. The framework is composed of an image library containing fruits and an improved faster \gls*{rcnn} model generation for better performance evaluation.

Different \gls*{dl} methods were revised in this section for the classification of fruits and identification of diseases. Nevertheless, no work was found to the best of our knowledge for the classification of strawberry runners or trusses. In this paper, semantic segmentation using both ResNet50 and ResNet101 are proposed to classify runners and trusses of strawberry plants for future pruning and harvesting applications that can extended to other crops.

\section{Methodology} \label{sec:imgsegmodel}
A custom dataset was generated for this project, which is now publicly available \cite{machado2022}.
The dataset is composed of 222 images, of which  194 images are for train and 28 images for testing. 57\% of the images contain only trusses, 12\% contain only runners, and the remaining 31\% of the images contain both trusses and runners.

The semantic segmentation was done using the PixelLib library\footnote{Available online, \protect\url{https://github.com/ayoolaolafenwa/PixelLib}, last accessed 17/06/2022} to train the model on the custom dataset. PixelLib offers the flexibility of training the model using ResNet50 and ResNet101 networks. In our work, we trained the model on both networks. Methodologies (such as Ga and Salt-and- artificially augment the dataset. The augmentation methods were used of the networks while training the model on the dataset. For each case, the training

Thirdly, evaluation; the evaluation is the \gls*{map}. Judging by the \gls*{map} one should be able to know if the model will be successful for inference with sample images. 

Lastly, inference; inference is the model's final component. This component is responsible for checking the performance of the model on new images that are not within the dataset. An image is inputted, and the model detects the trusses and runners within that image and then outputs a new image with the bounding boxes and labels drawn around the trusses and runners that have been predicted within the image.
The evaluation of this component is done using recall,  precision, and F1 score. These values are calculated according to equations \ref{eq:recall}, \ref{eq:precision} and \ref{eq:f1} respectively.

\begin{eqnarray}
\label{eq:recall}
Recall: re =\frac{tp}{tp+fn}
\end{eqnarray}

\begin{eqnarray}
\label{eq:precision}
Precision: pr =\frac{tp}{tp+fp}
\end{eqnarray}

\begin{eqnarray}
\label{eq:f1}
F1 score: f1 = 2\times \frac{pr.re}{pr+re}
\end{eqnarray}

Where, $tp$ is the true positives, $tn$ is the true negatives, $fp$ is the false positives and $fn$ is the false negatives.

\section{Results and Evaluations} \label{sec:resultsevaluations}

This section presents the result for the classification of trusses and runners.
Table \ref{tbl:evaluation} shows the results per network ResNet50, ResNet101 and per noise (i.e. Speckle, Salt-and-Pepper, Gaussian and Poison).

\begin{table}[h!]
\caption{Results from the evaluation component}

\begin{tabular}{lllll}
\hline
Architecture name & \gls*{map} (\%) & Number of epochs & Augmentation type & Parametrization  \\
\hline
ResNet50         & 1.2086\% & 300              & None              & None                 \\
ResNet50         & 1.3281\% & 300              & Speckle     & Severity = 2         \\
ResNet50         & 1.1368\% & 300              & Salt-and-Pepper   & 0.1                  \\
ResNet50         & 0.1333\% & 300              & Gaussian          & scale=0.2 x 255      \\
ResNet50         & 0.5\%    & 300              & Poisson           & 40                   \\
ResNet101        & 1.3168\% & 300              & None              & None                 \\
ResNet101        & 0.0625\% & 300              & Speckle     & Severity = 2         \\
ResNet101        & 0.4608\% & 300              & Salt-and-Pepper   & 0.1                  \\
ResNet101        & 1.0424\% & 300              & Gaussian          & scale=0.2 x 255      \\
ResNet101        & 1.1125\% & 300              & Poisson           & 40       \\
\hline
\end{tabular}
\label{tbl:evaluation}
\end{table}

\begin{figure}[h!]
	\centering
	\includegraphics[scale=0.5]{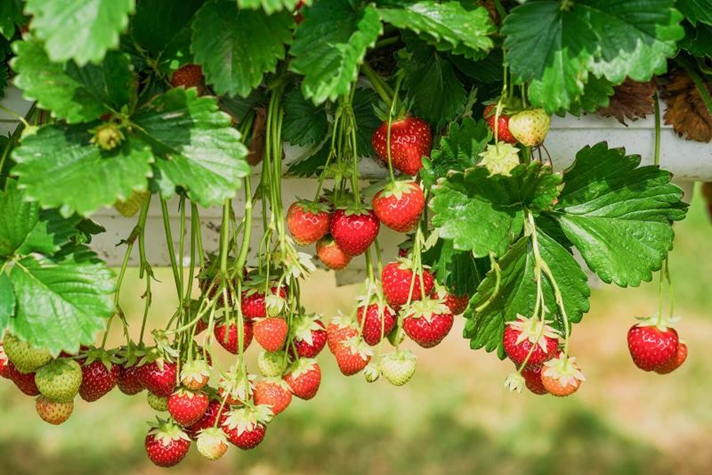}
	\caption{Input image for inference component}
	\label{fig:input}
\end{figure}

\begin{figure}[h!]
	\centering
	\includegraphics[scale=0.25]{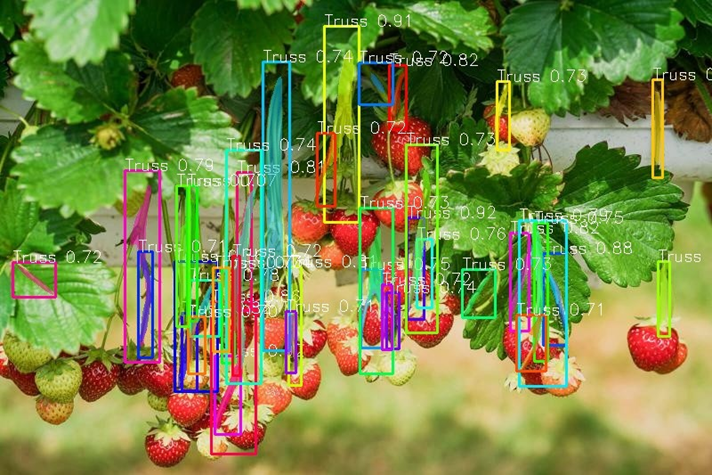}
	\includegraphics[scale=0.4]{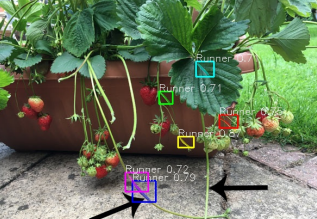}
	\caption{Example of classification. The left figure shows trusses being classified, while the right image shows also trusses and runners.}
	\label{fig:output}
\end{figure}

Inference used ResNet101 for backbone with 300 epochs and Salt-and-Pepper augmentation  added to 10\% of the images in the dataset.
The inference input and output image can be seen in figure \ref{fig:input} and figure \ref{fig:output}.

Table \ref{tbl:resultstrusses}, present the results for all models for trusses.
It is possible to observe that irrespective of the augmentation, ResNet101 results for trusses show high precision (values ranging from 0.88 to 0.91). The best result is for ResNet101 is obtained for Salt-and-Pepper augmentation, 0.91, 0.95 and 0.92 for precision, recall and F1 score (results highlighted in dark grey). 
ResNet50 with Poisson augmentation,  was able to successfully predict a large amount of trusses, having achieved 0.83, 0.53 and 0.65 for precision, recall and F1 score (results highlighted in light grey).

\begin{table}[h!]

\caption{Recall, precision and F1 scores for trusses}
\centering
\begin{tabular}{lllll}
\hline
Architecture name & Augmentation type & Precision & Recall & F1 score \\
\hline
ResNet50         & Speckle     & 0.05      & 0.01   & 0.02     \\
ResNet50         & Salt-and-Pepper   & 0         & 0      & 0        \\
ResNet50         & Gaussian          & 0.26      & 0.04   & 0.07     \\
\cellcolor[HTML]{dddddd} ResNet50         & \cellcolor[HTML]{dddddd} Poisson           & \cellcolor[HTML]{dddddd} 0.83      & \cellcolor[HTML]{dddddd} 0.53   & \cellcolor[HTML]{dddddd} 0.65     \\
ResNet101        & Speckle     & 0.88      & 0.83   & 0.85     \\
\cellcolor[HTML]{aaaaaa} ResNet101        & \cellcolor[HTML]{aaaaaa} Salt-and-Pepper   &\cellcolor[HTML]{aaaaaa}0.91      & \cellcolor[HTML]{aaaaaa} 0.95   & \cellcolor[HTML]{aaaaaa} 0.92     \\
ResNet101        & Gaussian          & 0.89      & 0.94   & 0.91     \\
ResNet101        & Poisson           & 0.89      & 0.95   & 0.91     \\
\hline
\end{tabular}
\label{tbl:resultstrusses}
\end{table}

Table \ref{tbl:resultsrunners}, present the results for all models for runners.
Opposed to trusses, runners always show a low value for precision (values ranging from 0 to 3\%). Similarly, this pattern can also be observed for recall and F1 score. 
As shown in table \ref{tbl:resultsrunners} no model using ResNet50 as backbone were able to predict any runners. Except ResNet101 combined with Salt-and-Pepper augmentation, the results are similar. The former had 3 runners successfully predicted, which could a consequence of only have runners 43\% of images. Furthermore, the model trusses and runners are very similar in shape and length and some runners were confused with trusses. 

\begin{table}[h!]

\caption{Recall, precision and F1 scores for runners}
\centering
\begin{tabular}{lllll}
\hline
Architecture name & Augmentation type & Precision & Recall & F1 score \\
\hline
ResNet50         & Speckle     & 0         & 0      & 0        \\
ResNet50         & Salt-and-Pepper   & 0         & 0      & 0        \\
ResNet50         & Gaussian          & 0         & 0      & 0        \\
ResNet50         & Poisson           & 0         & 0      & 0        \\
ResNet101        & Speckle     & 0         & 0      & 0        \\
ResNet101        & Salt-and-Pepper   & 0.03      & 0.01   & 0.02     \\
ResNet101        & Gaussian          & 0         & 0      & 0        \\
ResNet101        & Poisson           & 0         & 0      & 0        \\
\hline
\end{tabular}
\label{tbl:resultsrunners}
\end{table}

It is possible to observe that overall, ResNet101 outperformed ResNet50 when it comes to overall prediction of both trusses and runners. 

\section{Conclusions and Future Work} \label{sec:conlusionsfw}
The main objective of this paper is to develop a \gls*{ai} System that is capable to visually identify runners and trusses in strawberry plants. The dataset contains annotated images of both trusses and runners, trusses on their own, runners on their own and also very blurry images of all the previous.
It is also visible that the models struggle more with runners than trusses. ResNet101 results for trusses show high precision (values ranging from 88\% to 91\%) The best results for ResNet101 use Salt-and-Pepper augmentation, and achieved 91\%, 95\% and 92\% for precision, recall and F1 score; ; and 83\%, 53\% and 65\% on precision, recall and F1 score, respectively, for truss detection using the ResNet50 with dataset augmentation utilising Poisson noise. The results were expected because both Salt-and-Pepper and Poisson noises mimic the type of noise that cameras are exposed to in real and natural farming settings. 

Implementing different augmentations tried to enhance detection of runners, however  this not proved to be successful. There could be a number of different reason why the model struggles to classify runners. The difficulty resides in the fact that trusses and runners are very similar in size, shape, and colour.
A possible way to improve  results, one could try to apply different parameters to the image augmentations. A search space could be computed from combinations of the parameters, and using a meta-heuristic it would be possible to obtain better results. Alternatively, implementing other types of augmentation such as image rotation for training the models could potentially improve results. Future work also includes testing the model in a real scenario, where a pruning robot could use the model proposed in this work to automatically prune old trusses and runners. 

%
%
\bibliographystyle{unsrt}
\bibliography{references}

\end{document}

%% file: glossary.tex
\newacronym{cnn}{CNN}{Convolutional Neural Network}
\newacronym{dl}{DL}{Deep Learning}
\newacronym{ai}{AI}{Artificial Intelligence}
\newacronym{rcnn}{R-CNN}{Region-based Convolutional Neural Network}
\newacronym{lda}{LDA}{Latent Dirichlet Allocation}
\newacronym{msbslda}{MSBS-LDA}{Mean-Shift Bandwidths Searching Latent Dirichlet Allocation}
\newacronym{fcn}{FCN}{Faster Convolutional Network}
\newacronym{map}{MAP}{Mean Average Precision}
\newacronym{fps}{fps}{frames per second}

%% file: strawberry_paper.bbl
\begin{thebibliography}{10}

\bibitem{haenlein}
Michael Haenlein and Andreas Kaplan.
\newblock A brief history of artificial intelligence: On the past, present, and
  future of artificial intelligence.
\newblock {\em California management review}, 61(4):5--14, 2019.

\bibitem{bbc2021}
BBC.
\newblock Uk worker shortage: Farmers give fruit and veg awayfor free, 2021.
\newblock [Online; accessed 09/01/2021].

\bibitem{poling2012strawberry}
E~Barclay Poling.
\newblock Strawberry plant structure and growth habit.
\newblock {\em New York State Berry Growers Association, Berry EXPO}, 2012.

\bibitem{zheng2021remote}
Caiwang Zheng, Amr Abd-Elrahman, and Vance Whitaker.
\newblock Remote sensing and machine learning in crop phenotyping and
  management, with an emphasis on applications in strawberry farming.
\newblock {\em Remote Sensing}, 13(3):531, 2021.

\bibitem{behera2020}
Santi~Kumari Behera, Amiya~Kumar Rath, Abhijeet Mahapatra, and Prabira~Kumar
  Sethy.
\newblock Identification, classification \& grading of fruits using machine
  learning \& computer intelligence: a review.
\newblock {\em Journal of Ambient Intelligence and Humanized Computing}, pages
  1--11, 2020.

\bibitem{naik2017}
Sapan Naik and Bankim Patel.
\newblock Machine vision based fruit classification and grading-a review.
\newblock {\em International Journal of Computer Applications}, 170(9):22--34,
  2017.

\bibitem{bhargava2021}
Anuja Bhargava and Atul Bansal.
\newblock Fruits and vegetables quality evaluation using computer vision: A
  review.
\newblock {\em Journal of King Saud University-Computer and Information
  Sciences}, 33(3):243--257, 2021.

\bibitem{Bargoti2017}
Suchet Bargoti and James Underwood.
\newblock Deep fruit detection in orchards.
\newblock In {\em 2017 {IEEE} International Conference on Robotics and
  Automation ({ICRA})}. {IEEE}, may 2017.

\bibitem{Lamb2018}
Nikolas Lamb and Mooi~Choo Chuah.
\newblock A strawberry detection system using convolutional neural networks.
\newblock In {\em 2018 {IEEE} International Conference on Big Data (Big Data)}.
  {IEEE}, dec 2018.

\bibitem{wang2018unsupervised}
Yi~Wang and Lihong Xu.
\newblock Unsupervised segmentation of greenhouse plant images based on
  modified latent dirichlet allocation.
\newblock {\em PeerJ}, 6:e5036, 2018.

\bibitem{Kestur2019}
Ramesh Kestur, Avadesh Meduri, and Omkar Narasipura.
\newblock {MangoNet}: A deep semantic segmentation architecture for a method to
  detect and count mangoes in an open orchard.
\newblock {\em Engineering Applications of Artificial Intelligence}, 77:59--69,
  jan 2019.

\bibitem{tian2019}
Yunong Tian, Guodong Yang, Zhe Wang, Hao Wang, En~Li, and Zize Liang.
\newblock Apple detection during different growth stages in orchards using the
  improved yolo-v3 model.
\newblock {\em Computers and Electronics in Agriculture}, 157:417--426, 02
  2019.

\bibitem{Kang2020}
Hanwen Kang and Chao Chen.
\newblock Fruit detection, segmentation and 3d visualisation of environments in
  apple orchards.
\newblock {\em Computers and Electronics in Agriculture}, 171:105302, apr 2020.

\bibitem{ramdani2021strawberry}
Aldi Ramdani and Suyanto Suyanto.
\newblock Strawberry diseases identification from its leaf images using
  convolutional neural network.
\newblock In {\em 2021 IEEE International Conference on Industry 4.0,
  Artificial Intelligence, and Communications Technology (IAICT)}, pages
  186--190. IEEE, 2021.

\bibitem{boursianis2022internet}
Achilles~D Boursianis, Maria~S Papadopoulou, Panagiotis Diamantoulakis, Aglaia
  Liopa-Tsakalidi, Pantelis Barouchas, George Salahas, George Karagiannidis,
  Shaohua Wan, and Sotirios~K Goudos.
\newblock Internet of things (iot) and agricultural unmanned aerial vehicles
  (uavs) in smart farming: a comprehensive review.
\newblock {\em Internet of Things}, 18:100187, 2022.

\bibitem{machado2022}
Pedro Machado.
\newblock Strawberry dataset for semantic segmentation, June 2022.
\newblock https://doi.org/10.5281/zenodo.6656332.

\end{thebibliography}
